\documentclass[conference]{IEEEtran}
\usepackage{cite}
\usepackage{amsmath,amssymb,amsfonts}
\usepackage{algorithmic}
\usepackage{graphicx}
\usepackage{textcomp}
\usepackage{xcolor}
\usepackage{verbatim}
\usepackage{color,soul}
\usepackage{multirow}
\usepackage{balance}
\usepackage{url}
\def\BibTeX{{\rm B\kern-.05em{\sc i\kern-.025em b}\kern-.08em
    T\kern-.1667em\lower.7ex\hbox{E}\kern-.125emX}}
\begin{document}

\title{Operation-wise Attention Network for Tampering Localization Fusion}

\author{\IEEEauthorblockN{Polychronis Charitidis}
\IEEEauthorblockA{\textit{ITI-CERTH} \\
Thessaloniki, Greece \\
charitidis@iti.gr}
\and
\IEEEauthorblockN{Giorgos Kordopatis-Zilos}
\IEEEauthorblockA{\textit{ITI-CERTH} \\
Thessaloniki, Greece \\
georgekordopatis@iti.gr}
\and
\IEEEauthorblockN{Symeon Papadopoulos}
\IEEEauthorblockA{\textit{ITI-CERTH} \\
Thessaloniki, Greece \\
papadop@iti.gr}
\and
\IEEEauthorblockN{Ioannis Kompatsiaris}
\IEEEauthorblockA{\textit{ITI-CERTH} \\
Thessaloniki, Greece \\
ikom@iti.gr}
}

\maketitle

\begin{abstract}
In this work, we present a deep learning-based approach for image tampering localization fusion. This approach is designed to combine the outcomes of multiple image forensics algorithms and provides a fused tampering localization map, which requires no expert knowledge and is easier to interpret by end users. Our fusion framework includes a set of five individual tampering localization methods for splicing localization on JPEG images. The proposed deep learning fusion model is an adapted architecture, initially proposed for the image restoration task, that performs multiple operations in parallel, weighted by an attention mechanism to enable the selection of proper operations depending on the input signals. This weighting process can be very beneficial for cases where the input signal is very diverse, as in our case where the output signals of multiple image forensics algorithms are combined. Evaluation in three publicly available forensics datasets demonstrates that the performance of the proposed approach is competitive, outperforming the individual forensics techniques as well as another recently proposed fusion framework in the majority of cases.
\end{abstract}


\section{Introduction}
Image forensics algorithms are crucial for determining the integrity of digital images. This is evident by the number of research efforts in the literature that deal with different types of digital manipulations and traces \cite{korus2017digital}. Designing algorithms for detecting and localizing a specific type of forgery has proven to be challenging, even when evaluating their performance on controlled datasets and settings \cite{zampoglou2017large}. The situation is even more difficult when dealing with online media. Images are manipulated multiple times with different tools 
and are then circulated over the Internet, further undergoing a variety of transformations (e.g., cropping, re-sizing, re-compression). Consequently, many methods suffer from low detection accuracy and localization robustness, presenting noisy outcomes and higher false-positive rates when applied to new datasets \cite{zampoglou2017large}. Due to these observations, many works posit that there is true benefit in acquiring different reports from independent algorithms and inspecting the multiple clues in tandem \cite{fontani2013framework, iakovidou2020knowledge, kaur2019fusion, li2017image}. This is attributed to the fact that at least one or few of these diverse approaches are more likely to capture some tampering traces even after an image has undergone multiple transformations.

Although discovering manipulation traces is desirable, adding more forensics output visualizations increases the complexity of a media verification tool that presents the results, especially for non-expert users. The reason for this is that each algorithm has a different output that requires specific knowledge for proper interpretation. Consequently, this quickly becomes overwhelming for the non-expert. As a matter of fact, even forensics experts may provide wrong or contradicting interpretations of these visualizations \cite{spiegel_article}. 

In this paper, we aim to overcome these limitations, which often exclude non-expert users from the verification process of suspicious online images. The main objective is to develop a fully automatic fusion approach using deep learning, being able to leverage diverse forensics signals that are complementary to each other, so as to improve the robustness and reliability of the overall localization system.
This final visualization retains the most important features of the individual algorithms and discards the noise. Additionally, another important objective is to communicate the tampering localization results to end users in a manner that is easier to interpret and requires no additional specialized knowledge. This outcome can empower non-experts in image forensics, who are interested in image verification, e.g. fact-checkers and journalists, to actively contribute to image verification tasks.

The work presented in this paper is conducted in the context of the \textit{WeVerify} project \cite{marinova2020weverify}, which aims to build an open-source platform that engages communities and citizen journalists alongside newsroom and freelance journalists for collaborative and decentralized content verification, tracking, and debunking. For the content verification task, we build on the existing
\textit{Image Verification Assistant} \cite{zampoglou2016web}, which is a tool that analyses the authenticity of images by visualizing the outcomes of various tampering localization approaches. For the purpose of \textit{WeVerify}, we have extended this tool with new state-of-the-art algorithms but also by adding the presented fusion approach in order to enhance its usability by non-experts, including fact-checkers, journalists, and citizens. 



In this work, we make the following contributions. First, we present a fully automatic deep learning-based approach that learns how to combine diverse input signals into a more accurate overall tampering localization visualization. Second, the final output of our approach is easier to interpret and requires no specialized knowledge by end users. Finally, the evaluation results show that the proposed fusion approach largely leads to better results compared to other approaches.


\section{Related Work}
\label{related_work}

The field of media forensics has become increasingly popular in recent years \cite{verdoliva2020media}.
Several fusion approaches have been proposed in the literature, aiming at combining diverse signals that improve the overall robustness of the forensics localization maps. The various approaches can be categorized based on the level at which fusion is performed and the manipulation traces that are taken into consideration. Frameworks proposed for feature-level fusion often suffer from drawbacks related to selecting and handling a large number of features \cite{barni2012fuzzy,chetty2010nonintrusive}. Additionally, authors in \cite{iakovidou2020knowledge} present a framework that fuses the localization outputs of various forensics algorithms and further refines the result based on statistical features. The drawback of this method however is that it has hard-coded heuristics and assumptions about the input tampering localization algorithms.

Approaches based on measurement-level fusion are best suited for the tampering detection, and not localization, problem as they provide an overall result regarding forgeries being present or not \cite{fontani2013framework,kaur2019fusion}. 

On the other hand, techniques proposed for pixel-level fusion usually involve utilizing probability output maps and a fusion model to refine the final output and improve the localization of the tampered region \cite{li2017image}. In recent years, deep learning shows great power in many research fields, and methods based on deep Convolutional Neural Networks (CNN) outperform traditional methods and achieve remarkable success in solving computer vision problems. For instance, the work in  \cite{liu2018deep} presents a fusion architecture that outperforms traditional approaches. The difference of this approach is that it fuses features from different patches of the same image to produce the final localization map. 


\section{Methodology}
\label{methodology}

The main objective of our work is to develop a fully automatic fusion approach, able to exploit diverse forensics signals from the \textit{Image Verification Assistant} and generate a robust and easy-to-interpret visualization. The current version of \textit{Image Verification Assistant} contains 11 forensics algorithms with corresponding outputs. 

To simplify the training and evaluation of our fusion models, we select a subset of the forensics algorithms to be considered for fusion. Also, to provide a direct comparison with another fusion framework \cite{iakovidou2020knowledge}, we adopt their algorithm selection based on an evaluation of three publicly available datasets.
These datasets are: (i) The First IFS-TC Image Forensics Challenge set \cite{ieee_ifs}, (ii) the synthetic dataset by Fontani et al. \cite{fontani2013framework}, and (iii) the real cases of the Wild Web dataset \cite{zampoglou2015detecting}.
Based on the evaluation results, a set of five methods were selected as the building blocks of the fusion model:

\begin{itemize}
    \item ADQ1 \cite{lin2009fast} and DCT \cite{ye2007detecting} that both base their detection on analysis of the JPEG compression, in the transform domain; 
    \item BLK \cite{li2009passive} and CAGI \cite{iakovidou2018content} that base their detection on analysis of the JPEG compression in the spatial domain; 
    \item Splicebuster \cite{cozzolino2015splicebuster}, a noise-based detector selected as a complementary method due to its high reported performance and good interpretability of its produced outputs.
\end{itemize}
In this work, we adopt a deep learning-based fusion approach for the following reasons. First, we aspire to develop a fully automatic approach without the need for heuristic tuning or manual intervention, as for instance in the case of \cite{iakovidou2020knowledge}. Second,  the complex and diverse nature of the input signal calls for an effective approach to automatically extract the most important features, which deep learning excels at \cite{voulodimos2018deep}.
Finally, the availability of large-scale datasets, which are required by deep learning approaches makes the training of a deep learning-based model feasible.

For the fusion architecture, we initially consider two different deep learning architectures.
The first model we adopt is U-Net \cite{ronneberger2015u}. U-Net is a convolutional neural network that was initially applied for semantic segmentation in a medical context in order to localize tumors in the lungs or brain, but nowadays, it has got a much broader application field. The main idea is to supplement a usual contracting network by successive layers, where pooling operations are replaced by up-sampling operators. Hence these layers increase the resolution of the output. A successive convolutional layer can then learn to assemble a precise output based on this information. One important feature in U-Net is that there are a large number of feature channels in the up-sampling part, which allow the network to propagate context information to higher resolution layers. Consequently, the expansive path is roughly symmetric to the contracting part and yields a U-shaped architecture. The network only uses the valid part of each convolution without any fully connected layers. The U-Net architecture has a lot of variants. For the fusion task, we use \emph{Eff-Unet} \cite{baheti2020eff}, a variant of U-Net architecture that uses \emph{EfficientNet} \cite{tan2019efficientnet} as the encoder part. To adapt the \emph{Eff-Unet} to our fusion task, we use the \emph{EfficientNet-B4} variant as the encoder. We refer to this model as \emph{Eff-B4-Unet}.

\begin{figure*}[t]
\centerline{\includegraphics[width=0.6\textwidth]{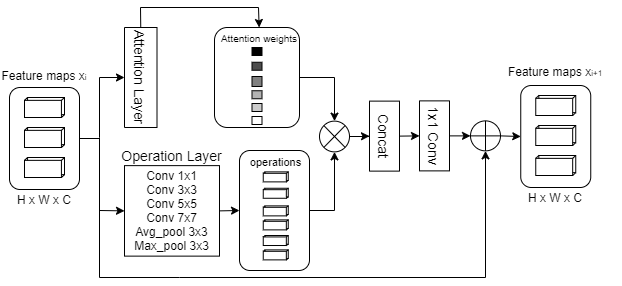}}
\caption{The architecture of adapted operation-wise attention layer, comprising an attention and operation layer, a concatenation operation, and 1 × 1 convolution.}
\label{fig:fusion}
\end{figure*}

The second model that we employ is a simple architecture of neural networks that was proposed for the problem of image restoration \cite{suganuma2019attention}. It performs multiple operations in parallel, which are weighted by an attention mechanism to enable the selection of proper operations depending on the input. The layer can be stacked to form a deep network, which is differentiable and thus can be trained in an end-to-end fashion by gradient descent. This architecture is suitable for our problem because it uses attention to capture important features by examining which operations are the most beneficial, depending on the input signal. Another important aspect of this architecture is that it can learn to attend low-level features, which is important for the fusion task, as semantic or high-level representations are often not relevant or useful for the problem. 
The network includes operations like, dilated convolutions, separable convolutions, and pooling with various kernel sizes and rates, which are weighted by an attention mechanism. To adapt this architecture to our needs, we replaced dilated convolutions with simple convolutions as the former are better suited for the image restoration task and exhibit lower validation performance in preliminary experiments.
Additionally, we reduce the number of stacked Operation-wise Attention (OwA) layers from 10 to 4, in order to avoid overfitting during training. Also, in our adaptation, each operation-wise attention layer consists of one operation layer, in contrast with the original implementation that stacks 4 consecutive operation layers. Figure \ref{fig:fusion} shows the modified operation-wise attention layer. In the operation layer, we use simple convolutions with filter sizes 1 × 1, 3 × 3, 5 × 5, 7 × 7, average and max pooling with a 3 × 3 receptive field. We refer to this model as \textit{OwAF}.

\section{Experimental study}
\label{experimental_study}

In this section, we compare these two deep learning approaches and we use the best performing in further evaluations, in order to compare it with another fusion approach against two publicly available image forensics datasets.

\subsection{Training and evaluation setup}
\label{training_and_evaluation_setup}

\noindent\paragraph{Training set} To train the deep learning models that we describe in Section \ref{methodology}, we need to use an image forensics dataset for training. For the model training in the fusion task, we generate a set of tampering localization outputs by applying the forensics algorithms for each image in the dataset.



Currently, there are not many large-scale forensics datasets available, and this is mainly due to the fact that their generation is very time-consuming. On the other hand, it is easy to generate synthetic forgeries in images. For this work, we use the DEFACTO dataset \cite{mahfoudi2019defacto}, which contains more than 150,000 images with synthetic forgeries and their corresponding ground truth masks. The dataset contains four manipulation types, including splicing, copy-move, object removal, and face morphing. For our task, we use only images from the first three manipulation types. We randomly compile 15,000 images for training, 1000 for test, and 1000 for validation. For every input image in each set, we produce a set of different tampering localization maps obtained by the selected detection methods. This means that the produced training set for the fusion model consists of $15,000 \times 5$ tampering localization maps and corresponding ground truth masks for every image. We concatenate the forensics maps in the channel axis to generate training samples.

\noindent\paragraph{Parameter selection} We use the Adam optimizer \cite{kingma2014adam} with an initial learning rate of 0.001, and we minimize the pixel-size binary cross-entropy loss. To achieve this, the final layer of both architectures has a sigmoid activation function. Ground truth masks are also in the [0,1] range, having zeros for the pristine parts of the image and ones for the tampered regions. The training batch size is set to 4. We train the model for 20 epochs, and we save only the weights of the model that has the lowest validation error. The validation error is derived from the evaluation of the images in the DEFACTO validation set. For faster training, we only use 300 images of the validation set. We also reduce the learning rate by a factor of 10 if the validation error does not improve for 5 epochs.

\noindent\paragraph{Datasets} To evaluate the models, we use three datasets. First, we evaluate the proposed deep learning models against the DEFACTO test dataset that contains 1000 images. Then, we choose the best of the two models and compare them to another previously proposed model \cite{iakovidou2020knowledge}.
This comparison is performed on two different forensics datasets. The first is the IFS-TC Image Forensics Challenge dataset \cite{ieee_ifs}, which contains 450 images with forgeries and is designed to serve as a realistic benchmark. Focusing on splicing tampering localization, we excluded cases that were produced by copy-move operations resulting in a set of 306 forgery cases produced through splicing operations only. Tampered images in this dataset are accompanied by ground-truth maps. The second dataset is the CASIA V2.0 dataset \cite{dong2013casia} that contains 5,123 realistically tampered color images of varying sizes. This dataset does not come with ground truth maps. In order for us to be able to perform localization tests, we take the ground truth masks that were produced in another work \cite{iakovidou2020knowledge}, where 2,195 reliable ground truth maps were manually produced through semi-automated procedures involving image differencing, thresholding, and morphological operations.

\noindent\paragraph{Evaluation metrics:} We use the F1 score, both the macro average and the F1 for the tampered class, and the average Intersection over Union (IoU) metric. 

\begin{figure*}[t]
\centerline{\includegraphics[width=0.65\textwidth]{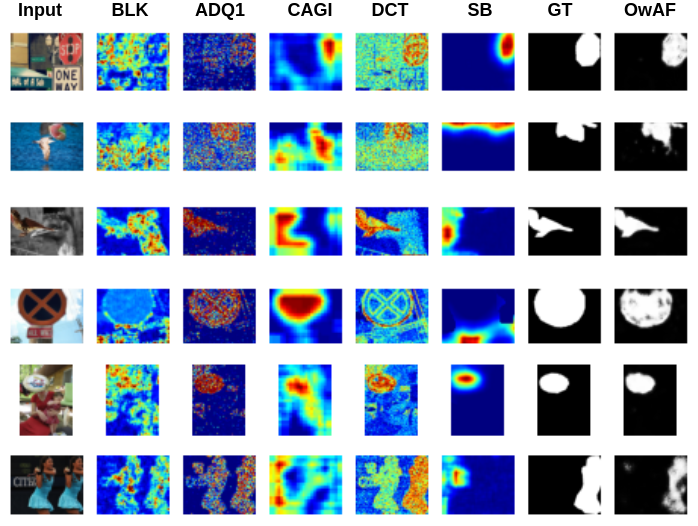}}
\caption{Various examples of tampering localization outputs using input images from the DEFACTO test dataset and fusion visualizations from the \textit{OwAF} model; GT column refers to ground truth masks. }
\label{fig:defacto_results}
\end{figure*}

\subsection{Results}
\label{results}

In the first experiment, we investigate the performance of the two proposed fusion models. Table \ref{tab:defacto_eval} shows the evaluation results of the two proposed models in the DEFACTO test dataset as well as results from all individual algorithms.

\begin{table}[htbp]
\caption{Evaluation metrics on DEFACTO test dataset for the proposed fusion models and the individual localization algorithms. The binarization threshold is set to 0.5}
\begin{center}
\begin{tabular}{|c|c|c|c|}
\hline
\textbf{Models} & \textbf{Macro-F1} & \textbf{F1 (tampered)} & \textbf{IoU}   \\ \hline \hline
BLK \cite{li2009passive}             & 0.463             & 0.090                  & 0.053          \\ \hline
ADQ1 \cite{lin2009fast}           & 0.573             & 0.221                  & 0.123          \\ \hline
CAGI \cite{iakovidou2018content}           & 0.479             & 0.142                  & 0.072          \\ \hline
DCT \cite{ye2007detecting}         & 0.509             & 0.186                  & 0.101          \\ \hline
Splicebuster \cite{cozzolino2015splicebuster}    & 0.554             & 0.153                  & 0.087          \\ \hline
Eff-B4-Unet     & 0.908             & 0.798                  & 0.690          \\ \hline
OwAF     & \textbf{0.912}    & \textbf{0.829}         & \textbf{0.707} \\ \hline
\end{tabular}
\label{tab:defacto_eval}
\end{center}
\end{table}

\begin{figure*}[t]
\centerline{\includegraphics[width=0.67\textwidth]{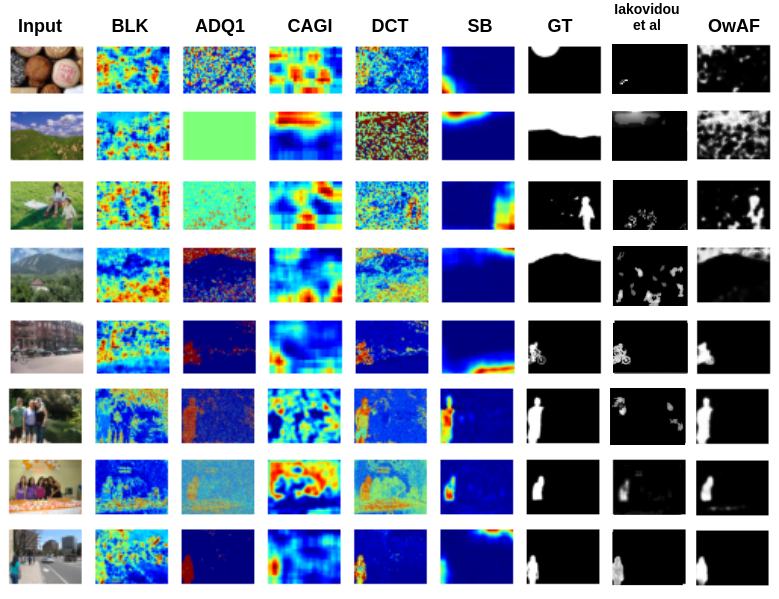}}
\caption{Successful examples of tampering localization outputs from the \textit{OwAF} model and visual comparisons with the individual methods and Iakovidou et al. \cite{iakovidou2020knowledge}; for the first four rows, images are taken from the CASIA v2 dataset, for the last four rows from the First IFS-TC Image Forensics Challenge dataset.}
\label{fig:eval_results}
\end{figure*}

We can see that the \textit{OwAF} network outperforms the \textit{Eff-B4-Unet} in every evaluation metric. Evaluation results for individual algorithms are very low when compared to the fusion approaches. This can be attributed to the fact that certain algorithms work well on specific manipulation types and images. Also, the 0.5 binarization threshold may impair the evaluation performance of many algorithms, as many algorithms might operate on lower thresholds. Averaging the results from many binarization ranges did not significantly improve the evaluation results, nonetheless. The best performing individual model is ADQ1. Another important observation is that the macro-averaged F1 score is higher compared to the F1 score for the tampered class for all evaluations. This means that the models can predict the pristine regions more accurately than the tampered ones. In terms of intersection over union (IoU) evaluation, we can see that the \textit{OwAF} model still outperforms \textit{Eff-B4-Unet}. 

Figure \ref{fig:defacto_results} shows random examples from the DEFACTO test dataset. The first column includes the input images. The next five columns show the outputs of the individual tampering localization algorithms. The final two columns show the ground truth mask, which reveals the actual location of the forgery and the fusion result of the \textit{OwAF} model that combines the localization outputs, respectively. It is evident from these examples that the fusion architecture learned to combine the diverse signals in order to localize the tampered region. One interesting observation is that for each input example, there are usually different algorithms that better localize the forgery. It seems that the fusion model learned to detect proper signals that contribute to a correct localization. For example, in the first row, Splicebuster and CAGI spot the tampering, but in row three, ADQ1 and DCT do so. In both cases, the fusion model has identified these signals and provides a correct result. 

To further investigate the fusion performance, we compare our best performing approach with another fusion framework proposed by \cite{iakovidou2020knowledge}. For evaluation, we use the two datasets described above (CASIA v2 and IFS-TC) in order to examine the generalization capabilities of the fusion model that was trained with the DEFACTO dataset. CASIA v2 is the first evaluation dataset we use. 
Table \ref{tab:eval} shows the evaluation results. The binarization threshold is set to 0.5.


\begin{table}[htbp]
\caption{Evaluation results of the CASIA v2 and First IFS-TC Image Forensics Challenge datasets. The binarization threshold for the metric calculation is set to 0.5}
\begin{center}
\begin{tabular}{|c|c|c|c|c|}
\hline
\textbf{Dataset}                                                     & \textbf{Models}         & \textbf{Macro-F1} & \textbf{F1 (tampered)} & \textbf{IoU}   \\ \hline \hline
\multirow{7}{*}{\begin{tabular}[c]{@{}c@{}}CASIA\\  v2\end{tabular}} & BLK  \cite{li2009passive}                    & 0.509             & 0.140                  & 0.089          \\ \cline{2-5} 
                                                                     & ADQ1  \cite{lin2009fast}                    & 0.573             & 0.217                  & 0.130          \\ \cline{2-5} 
                                                                     & CAGI  \cite{iakovidou2018content}                  & 0.502             & 0.162                  & 0.094          \\ \cline{2-5} 
                                                                     & DCT   \cite{ye2007detecting}                   & 0.546             & 0.196                  & 0.113          \\ \cline{2-5} 
                                                                     & Splicebuster   \cite{cozzolino2015splicebuster}          & 0.576             & 0.163                  & 0.093          \\ \cline{2-5} 
                                                                     & Iakovidou et al. \cite{iakovidou2020knowledge} & 0.598             & 0.253                  & 0.166          \\ \cline{2-5} 
                                                                     & OwAF             & \textbf{0.611}    & \textbf{0.270}         & \textbf{0.172} \\ \hline \hline
\multirow{7}{*}{IFS-TC}                                              & BLK   \cite{li2009passive}                   & 0.459             & 0.092                  & 0.063          \\ \cline{2-5} 
                                                                     & ADQ1  \cite{lin2009fast}                    & 0.485             & 0.115                  & 0.076          \\ \cline{2-5} 
                                                                     & CAGI    \cite{iakovidou2018content}                & 0.506             & 0.135                  & 0.091          \\ \cline{2-5} 
                                                                     & DCT     \cite{ye2007detecting}                 & 0.467             & 0.106                  & 0.065          \\ \cline{2-5} 
                                                                     & Splicebuster \cite{cozzolino2015splicebuster}            & \textbf{0.560}    & \textbf{0.199}         & \textbf{0.129} \\ \cline{2-5} 
                                                                     & Iakovidou et al. \cite{iakovidou2020knowledge} & 0.549             & 0.171                  & 0.112          \\ \cline{2-5} 
                                                                     & OwAF             & 0.529             & 0.140                  & 0.106          \\ \hline
\end{tabular}
\label{tab:eval}
\end{center}
\end{table}

We can observe slightly better performance in every metric from individual models compared to those in Table \ref{tab:defacto_eval}. This means that this dataset contains images with manipulations that can be localized better by the individual algorithms. ADQ1 and DCT are still the best performing individual approaches. Regarding the fusion methods evaluation, we can see that our approach outperforms the competing fusion framework \cite{iakovidou2020knowledge}. One notable observation is that the performance of OwAF is significantly worse than the evaluation results that are reported in Table \ref{tab:defacto_eval}. This is a clear indication that our trained models have overfitted to the training set manipulations. The fusion model possibly learned to localize specific forgeries, like shapes and patterns from the outputs of individual algorithms that frequently appear in the DEFACTO dataset. For example, the DEFACTO dataset contains synthetic manipulations on specific objects such as traffic signs, birds, airplanes, humans (Figure \ref{fig:defacto_results}). Yet, the proposed approach is still better than individual algorithms and also outperforms the fusion framework of \cite{iakovidou2020knowledge} in terms of both F1 score and IoU.

Evaluation results on the First IFS-TC Image Forensics Challenge dataset are also reported in Table \ref{tab:eval}. In these results, a significant decrease in the performance of the individual algorithms can be observed. One exception is the Spicebuster performance, which increased compared to previous evaluations. Splicebuster even outperforms both fusion approaches. Iakovidou et al. \cite{iakovidou2020knowledge} also achieve marginally better performance than our fusion model in this dataset. One possible explanation is that our fusion model learned to focus more on the individual localization maps that achieved better performance in the training set, namely the ADQ1 and DCT. This can also be verified from the CASIA v2 evaluation results, where ADQ1 and DCT are still the best individual approaches, and our approach outperforms all others. On the contrary, in the First IFS-TC Image Forensics Challenge case, the best performing individual algorithm is Splicebuster and this possibly justifies the poor performance of \textit{OwAF}.

Figure \ref{fig:eval_results} shows some successful examples of tampering localization outputs produced by the fusion model and the individual methods from the two evaluation datasets. In most examples, ADQ1 and DCT visualization better localize the tampering. There are also cases where Splicebuster and CAGI better localize the forgeries. Additionally, in these specific examples, we can see that the visual result from the method of Iakovidou et al. \cite{iakovidou2020knowledge} is very poor for the CASIA v2 samples (first four). 
The qualitative results are improved for the IFS-TC samples (last four), which is expected as this approach overall performs better in this dataset. 

From our experiments, it is evident that the main challenge of the proposed approach stems from overfitting to the training data. Although \textit{OwAF} can outperform individual forensics algorithms or other fusion models, we observe a lack of generalization to unseen manipulations. Namely, we get poor predictions for datasets that have different types of manipulations compared to those that appeared in the training dataset. Additionally, the low evaluation performance of individual models is a major indication that the forgery localization problem is very difficult and is even more challenging to design a general fusion solution for images in the wild. These observations are also reported by other researchers \cite{zampoglou2017large}.



\section{Conclusions}
\label{conclusion}

The main objective of our work has been to develop a fully automatic fusion approach, able to exploit diverse signals from various forensics algorithms. The fusion outcome would communicate the tampering localization results to end users in a manner that is easier to interpret and requires no additional specialized knowledge. 

The reported experimental results are promising and in many cases outperform the individual forensics techniques and a competing fusion approach. An important limitation of this work is the generalization ability of the fusion model and stems from the general limitations of the underlying supervised learning scheme and the insufficient representativeness of the dataset that was used for training.

In the future, we plan to work on these generalization limitations. To this end, we will try to increase the size of the training dataset and include different manipulations from other synthetic datasets. Additionally, we will experiment with task-specific regularization techniques, like feature map dropout. Finally, another interesting research direction is to experiment with multi-stream fusion architectures that besides the forensics localization maps, will extract features from the input image itself.

\small
\section*{Acknowledgments}

This work has been supported by the WeVerify H2020 project, partially funded by the European Commission under contract number 825297. 
\normalsize

\balance
\bibliographystyle{IEEEtran}
\bibliography{references}

\end{document}